\documentclass{article}

\usepackage{microtype}
\usepackage{graphicx}
\usepackage{subcaption}
\usepackage{booktabs} 

\usepackage{hyperref}

\usepackage[preprint]{icml2026}

\usepackage[algo2e]{algorithm2e}
\usepackage{amsmath}
\usepackage{amssymb}
\usepackage{mathtools}
\usepackage{amsthm}

\usepackage{booktabs}
\usepackage{multirow}
\usepackage{colortbl}
\definecolor{table-blue}{RGB}{173, 216, 230}
\usepackage{xcolor}
\usepackage{soul}
\usepackage[capitalize,noabbrev]{cleveref}
\usepackage{xspace}
\usepackage{tcolorbox}

\usepackage{xcolor}
\usepackage{tcolorbox}
\usepackage{listings}

\tcbuselibrary{listings,skins,breakable}

\theoremstyle{plain}

\theoremstyle{definition}

\theoremstyle{remark}

\newcommand{\our}{\textsc{CHARM}\xspace}
\newcommand{\newparagraph}{\noindent\textbf}

\usepackage[textsize=tiny]{todonotes}

\usepackage{amsmath,amsfonts,bm}

\def\eqref#1{equation~\ref{#1}}

\def\1{\bm{1}}

\DeclareMathAlphabet{\mathsfit}{\encodingdefault}{\sfdefault}{m}{sl}
\SetMathAlphabet{\mathsfit}{bold}{\encodingdefault}{\sfdefault}{bx}{n}

\icmltitlerunning{Preprint. Under review.}

\begin{document}
\twocolumn[
  \icmltitle{\our: Calibrating Reward Models With Chatbot Arena Scores}



  \icmlsetsymbol{equal}{*}

  \begin{icmlauthorlist}
    \icmlauthor{Xiao Zhu}{equal,HKUST(GZ)}
    \icmlauthor{Chenmien Tan}{equal,Alibaba}
    \icmlauthor{Pinzhen Chen}{Queen}
    \icmlauthor{Rico Sennrich}{UZH}
    \icmlauthor{Huiming Wang}{SUTD}
    \icmlauthor{Yanlin Zhang}{HKUST(GZ)}
    \icmlauthor{Hanxu Hu}{UZH}

  \end{icmlauthorlist}

  \icmlaffiliation{HKUST(GZ)}{HKUST(Guangzhou)}
  \icmlaffiliation{Alibaba}{Alibaba Group}
  \icmlaffiliation{Queen}{Queen’s University Belfast}
  \icmlaffiliation{UZH}{University of Zurich}
  \icmlaffiliation{SUTD}{Singapore University of Technology and Design}


  \icmlcorrespondingauthor{Hanxu Hu}{hanxu.hu@uzh.ch}


  \vskip 0.3in
]



\printAffiliationsAndNotice{}  

\begin{abstract}
  Reward models (RMs) play a crucial role in Reinforcement Learning from Human Feedback by serving as proxies for human preferences in aligning large language models. However, they suffer from various biases which could lead to reward hacking. In this paper, we identify a model preference bias in RMs, where they systematically assign disproportionately high scores to responses from certain policy models, leading to unfair judgments. To mitigate this bias, we propose a calibration method named \textbf{CH}atbot \textbf{A}rena calibrated \textbf{R}eward \textbf{M}odeling (\our) that leverages Elo scores from the Chatbot Arena to construct debiased preference datasets and adjust reward model scoring. We conduct extensive experiments on reward model benchmarks and human preference alignment. Results demonstrate that our calibrated RMs achieve improved evaluation accuracy on RM-Bench and the Chat-Hard domain of RewardBench, exhibit a stronger correlation with human preferences by producing scores more closely aligned with Elo rankings and improve downstream post-training performance. These results demonstrate that \our provides a simple, effective, and broadly applicable approach to building more reliable and fair reward models. Our code is available at \href{https://github.com/HexagonStar/CHARM}{\texttt{https://github.com/HexagonStar/CHARM}}.
\end{abstract}

\section{Introduction}

Reinforcement Learning from Human Feedback \citep[RLHF;][]{ouyang2022training, christiano2017deep} has emerged as a fundamental approach for aligning large language models (LLMs) with human values, ensuring they generate helpful, coherent, and safe responses \citep{achiam2023gpt,touvron2023llama,team2023gemini,qwen}. At the core of RLHF are reward models (RMs). RMs are typically trained on pairwise preference data, where human annotators evaluate multiple model-generated responses and rank them based on specific criteria \citep{ouyang2022training, lee2024rlaif}. Given these ranked preferences, the RM learns to predict which responses humans would favor, effectively acting as an automated judge in place of human raters.

However, reward models are not infallible; their inherent biases can compromise the fairness of evaluations and allow policy models to exploit these vulnerabilities through reward hacking \citep{skalse2022defining}. In such scenarios, language models optimize their outputs to maximize the reward, often in ways that deviate from genuine human preferences \citep{eisenstein2023helping,gao2023scaling,pang2022reward}. Recent studies have revealed several such biases, which often manifest as a preference for specific content patterns, such as length bias \citep{huang2025posthoc} or style bias \citep{zhang2024lists}. The existence of these biases means a model can achieve a high score simply by generating longer or more elaborately styled text rather than by genuinely improving the reliability of its content, potentially leading to deceptive or unintended outcomes.

In this paper, our contributions can be summarized as follows:
\begin{itemize}
    \item We analyze the score distributions of various reward models across a wide range of policy models and identify a different and more subtle type of bias \textbf{Model Preference Bias}, as it manifests in reward models systematically giving disproportionately high scores to certain policy models, beyond what is justified by human preferences.
    \item We propose a calibration method named \textbf{CH}atbot \textbf{A}rena calibrated \textbf{R}eward \textbf{M}odeling (\our), which utilizes these over-valued policy models as generators of false positives and constructs debiased preference pairs by leveraging Elo scores from Chatbot Arena.
    \item Through extensive experiments, we demonstrate that \our consistently improves reward model performance in accordance with their initial bias. In particular, \our improves Skywork-Reward-Llama-3.1-8B-v0.2 by 5.2 points on RM-Bench Chat and by 1.6 points on average. Beyond reward modeling, \our also yields gains in downstream post-training. Applying \our-calibrated reward models improves Qwen2.5-7B-Instruct by 2.4 points on IFEval.
    \item Further analysis shows that \our makes reward models more robust to stylistic variations in model responses, and can be interpreted as a form of implicit pattern calibration.
\end{itemize}


\begin{figure*}[t]
    \centering
    \includegraphics[width=0.75\linewidth]{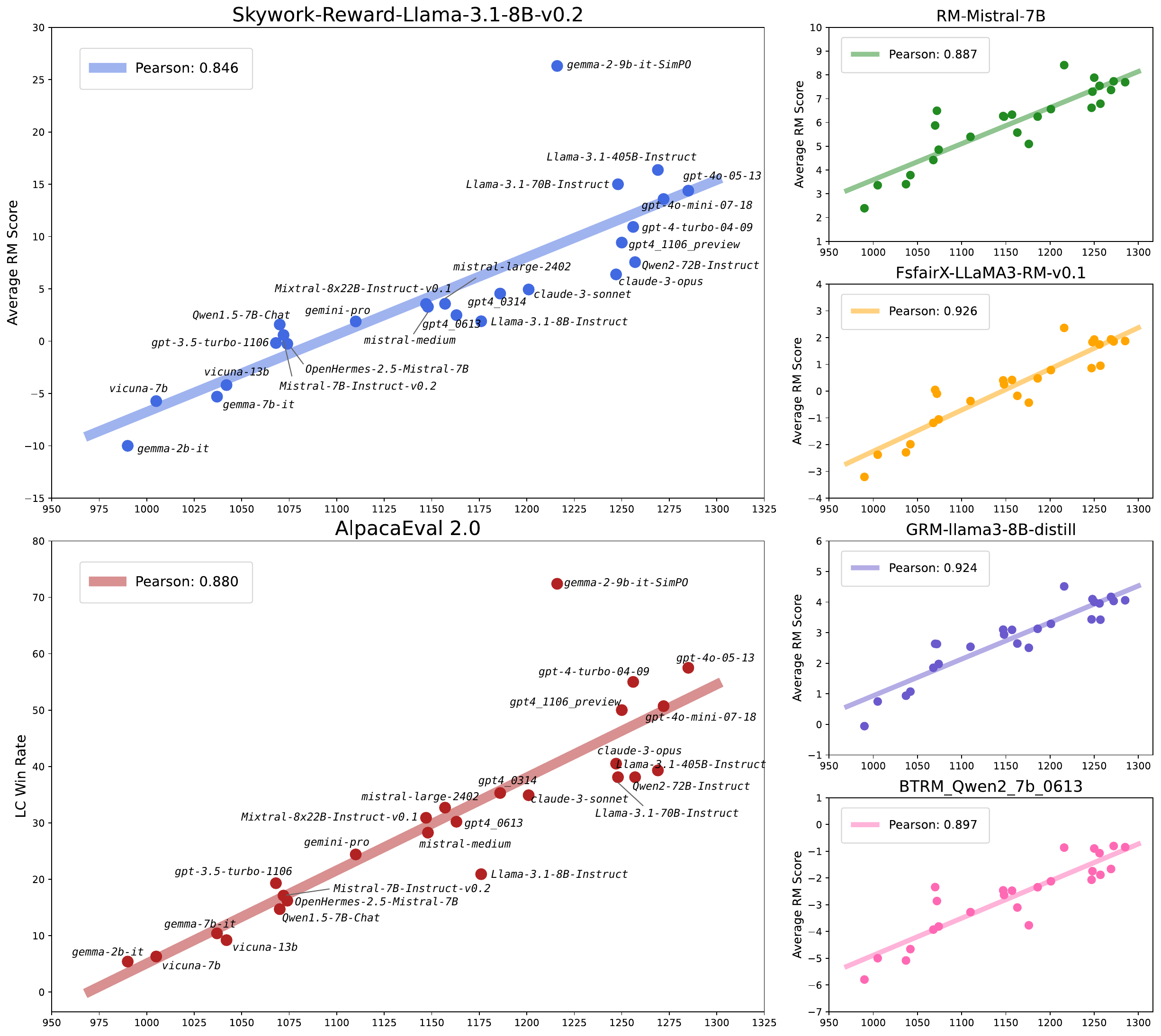}
    \caption{Average reward model scores across policy models on AlpacaEval. The x-axis represents arena Elo scores. The left lower plot illustrates the Length-Controlled win rates of these models on AlpacaEval. }
    \label{Alignment}
    \vspace{-0.45cm}
\end{figure*}

\section{Related Work}
\newparagraph{Reward Models} A reward model assigns scores to responses generated by large language models, helping rank and select the most human-aligned outputs \citep{ouyang2022training}. Formally, let $\mathcal{D}=\{(x,y)\}$ represent a dataset of instruction-response pairs, where $x\in\mathcal X$ is an instruction and $y\in\mathcal Y$ is a response. A reward model $r_\phi:\mathcal X\times\mathcal Y\to\mathbb R$ predicts the score $r_\phi(x,y)$ for a response $y$ conditioned on an instruction $x$. Most RMs are trained using pairwise preference data, which consists of triplets $(x, y^+, y^-)$, where $y^+$ is the preferred response over $y^-$. The RM is trained to optimize a Bradley-Terry pairwise ranking loss:
\begin{equation}
    \mathcal{L}(\phi) = -\mathbb E_{(x,y^+,y^-)\sim\mathcal D}\Big[\log \sigma\big(r_\phi(x,y^+) - r_\phi(x,y^-)\big)\Big]
\end{equation}

\newparagraph{LLM Evaluation.} Both automated benchmarks and human evaluation are widely used in LLM evaluation. A series of benchmarks such as MT-Bench \citep{zheng2023judging}, Alpaca-Eval \citep{dubois2023alpacafarm}, and Arena-hard \citep{li2024crowdsourceddatahighqualitybenchmarks} employ LLM-as-a-judge systems to assess the quality of model responses on a fixed set of prompts. A series of works, such as Ultrafeedback \citep{cui2024ultrafeedback} and RLAIF \citep{lee2024rlaif}, use LLMs for preference annotation which correlate training signals with evaluation measures. On the other side, ChatBot Arena \citep{chiang2024chatbot} employs a crowdsourced, pairwise comparison system where users challenge two anonymous models with prompts and select their preferred response. This vast collection of human preference data is then aggregated to compute a dynamic Elo rating for each model, which serves as a widely recognized proxy for genuine human preference.

\newparagraph{Bias Mitigation.} Reward models and LLM-as-a-judge might bring biases and affect the accuracy of the evaluation. \citet{park2024offsetbias} identified six distinct types of bias in evaluation models and leveraged LLMs to construct a debiased dataset. Beyond these biases, \citet{li2025preference} found that judge models may develop bias, favoring content generated by themselves or closely related LLMs due to their exposure to synthetic data. \citet{dubois2024lengthcontrolled} proposed a regression-based method to mitigate length bias, while \citet{huang2025posthoc} introduced a post hoc calibration technique for reward models. 



\section{Model Preference Bias: An Empirical Study}\label{PreferenceBiasinRewardModels}




Reward models are designed with the primary goal of aligning model responses with human preferences. One of the most direct reflections of human preferences in large-scale AI evaluation is Chatbot Arena \citep{zheng2023judging}, where real users interact with language models and rank them based on their responses. Given that Chatbot Arena represents a universal distribution of real-world prompts, we can make the assumption that an ideal RM should produce scores that strongly correlate with the platform's Elo ratings.

To validate this, we use AlpacaEval \citep{alpaca_eval} as our evaluation set, which consists of 805 carefully curated questions. This dataset has been shown to exhibit a 98\% Spearman correlation with Chatbot Arena. We evaluated five popular RMs \citep{liu2024skywork,dong2023raft,xiong2024iterative,yang2024regularizing} on a diverse set of policy models with varying Elo scores (Since some models on AlpacaEval did not participate in the Chatbot Arena, their Elo scores are unavailable. More results and statistics are in Appendix~\ref{appendix_preference_bias} and Appendix~\ref{appendix_elo_stat}). For each policy model, we calculated the average score from each RM. We also display the Length-Controlled win rates of these policy models on AlpacaEval \citep{dubois2024lengthcontrolled}. The results are shown in Figure~\ref{Alignment}:

\newparagraph{Observation 1: RM Scores Correlate Positively with Human Preferences.} From Figure~\ref{Alignment}, we observe that models with higher Elo scores in Chatbot Arena tend to receive higher RM scores on their responses. This supports our initial assumption that an ideal RM should reflect human preference rankings. We compute the Pearson correlation between policy models' RM and Elo scores for each RM (Also see Figure~\ref{Alignment}). The results indicate a strong positive correlation across all tested RMs.

\newparagraph{Observation 2: RMs May Favor Certain Policy Models Unfairly.} Although RM scores exhibit an overall alignment with human preferences, they sometimes deviate for specific policy models, assigning scores that are inconsistent with the models' Elo scores. Some policy models, such as Gemma-2-9b-it-SimPO \citep{meng2024simpo}, receive disproportionately high RM scores, sometimes even surpassing significantly stronger models based on Elo rankings. A similar trend is observed in the AlpacaEval leaderboard, suggesting that this bias may also exist in LLM-as-a-judge systems. Notably, the over-valued models share a similarity in that they undergo preference optimization, implying a potential source of systematic biases.


Based on these observations, we define this issue as \textbf{Model Preference Bias}, which occurs when reward models systematically assign unjustifiably high scores to certain policy models. This overestimation makes RMs fail to provide fair evaluations. To address this bias, we detail our proposed calibration method in the next section.

\section{Methodology}


\subsection{CHARM: Chatbot Arena Calibrated Reward Modeling}
Given a set of instructions $\mathcal{X} = \{x_i\}^N_{i=1}$ and two policy models, one over-valued model $\pi_O$ and one reference model $\pi_R$, for each instruction $x_i$, the two models generate responses $y_i^O \sim \pi_O$ and $y_i^R \sim \pi_R$. The reward model $r_\phi$ assigns scores to these responses, producing $s_i^O = r_\phi(x_i, y_i^O)$ and $s_i^R = r_\phi(x_i, y_i^R)$, resulting in a preference dataset $\mathcal{D}=\{(x_i,y_i^+,y_i^-)\mid y_i^+=\arg\max{(s_i^{O},s_i^{R})}\}^N_{i=1}$. Since the reward model $r_\phi$ overestimates the responses from model $\pi_O$, the resulting preference dataset $\mathcal{D}$ inherits this bias. To address this issue, \our reconstructs a debiased preference dataset to mitigate the preference bias in reward modeling.

The Elo rating system \citep{elo1967proposed} provides a probabilistic model for ranking players (or models, in this case) based on their relative performance. Following Chatbot Arena's implementation where a model gets a score of 1 for a win, 0.5 for a tie, and 0 for a loss, for an over-valued model with $\text{Elo}_{O}$ and a reference model with $\text{Elo}_{R}$, the expected win rate of the over-valued model $\mathbb{P}(O)$ is defined as Equation~\ref{eq:elo_definition}. $\mathbb{P}(O)$ can also be expressed as a weighted sum of the probabilities of the over-valued model getting a win $\mathbb{P}_{win}$ and a tie $\mathbb{P}_{tie}$:
\begin{align}
    \mathbb{P}(O) &= \frac{1}{1 + 10^{(\text{Elo}_R - \text{Elo}_O) / 400}} = \mathbb{P}_{win} + 0.5\mathbb{P}_{tie}
\label{eq:elo_definition}
\end{align}

Ties in RM are rare unless both models produce identical responses, naturally requiring $\mathbb{P}_{tie}\to0$. Nonetheless, Chatbot Arena's scoring implementation (1/0.5/0 for win/tie/loss) allows us to evenly split $\mathbb{P}_{tie}$ between wins and losses, maintaining equivalent Elo scores. Therefore, our Elo-derived win rate can be directly applicable to a strict win/loss scenario for RMs. Given the observation in section~\ref{PreferenceBiasinRewardModels} that RM scores are correlated with Elo scores, we contend that if the RM were perfectly aligned with human preferences, its empirical win rate should match this probability:
\begin{equation}\label{eq:calibrated_winrate}
\hat{\mathbb{P}}(O)=\frac{1}{N}\sum^N_{i=1}\sigma(s_i^O-s_i^R) \approx \mathbb{P}(O)    
\end{equation}

However, in practice, we find that there exist deviations between $\hat{\mathbb{P}}(O)$ and $\mathbb{P}(O)$ because of model preference bias. To correct this bias, we seek a transformation of RM scores such that the empirical win rate $\hat{\mathbb{P}}^\prime(O)$ after calibration better aligns with the expected win probability $\mathbb{P}(O)$. We introduce a score offset $\Delta$ applied to the RM scores of over-valued policy model's responses: $s_i^{\prime O}=s_i^O+\Delta$, then the calibrated empirical win rate will be: $\hat{\mathbb{P}}^\prime(O)=\frac{1}{N}\sum^N_{i=1}\sigma(s_i^{\prime O}-s_i^{R})$. Our goal is to find a $\Delta$ that minimizes the deviation from the theoretical probability. We optimize $\Delta$ by minimizing the MSE loss:

\begin{equation}\label{eq:loss}
    \mathcal{L}(\Delta)= \mathbf{MSE}\big(\frac{1}{N}\sum^N_{i=1}\sigma(s_i^{O} + \Delta - s_i^R),\mathbb{P}(O)\big)
\end{equation}

After determining the offset $\Delta$, we can construct a calibrated preference dataset $\mathcal{D}^\prime=\{(x_i,y_i^+,y_i^-)\mid y_i^+=\arg\max{(s_i^{O}+\Delta,s_i^{R})}\}^N_{i=1}$ for further reward model training. 








\subsection{A Metric for Model Preference Bias Measurement}

To quantify the misalignment between a reward model's preference for different policy models and human preferences, we introduce a Mismatch Degree metric. This metric shows the discrepancy between the reward model's scoring and the expected human preference reflected by Elo scores, measuring the degree of RM's model preference bias.

Given a model $\pi_O$, a reference model $\pi_R$, and a preference dataset built upon them, we define the Mismatch Degree (MD) between them as:
\begin{equation}
\textbf{MD}(\pi_{O},\pi_{R}) = \Bigg|\frac{\hat{\mathbb{P}}(O) - \mathbb{P}(O)}{\max(\mathbb{P}(O),1-\mathbb{P}(O))} \Bigg|
\label{md}
\end{equation}

where $\hat{\mathbb{P}}(O)$ is the probability of model $\pi_O$ winning against $\pi_R$ according to the reward model's scores. $\mathbb{P}(O)$ is the expected win rate of $\pi_O$ over $\pi_R$, derived from their Elo scores in Chatbot Arena. This metric captures how much the reward model's judgments deviate from the expected human preference. A positive $\hat{\mathbb{P}}(O) - \mathbb{P}(O)$ indicates that the reward model over-values model $\pi_O$ relative to what is expected from human preferences while a negative value indicates an under-value.

\section{Experiments}
In this section, we aim to address the following questions through experiments to validate the effectiveness of \our:

\newparagraph{Question 1.} Does \our enhance the reward model's judging capability, leading to more accurate and reliable evaluations? 

$\bullet$ We evaluate calibrated reward models on benchmarks such as RM-Bench and RewardBench, which consist of diverse instructions paired with two candidate responses. The reward model must assess and select the better response, providing a robust framework to measure its judging capability. 

\newparagraph{Question 2.} Does \our successfully reduce model preference bias, improving alignment with human preferences?

$\bullet$ We construct a battlefield using responses from various LLMs on AlpacaEval. Each response is scored by the reward models, allowing us to compute pairwise win rates between models. We then compare these RM-derived win rates against Elo-derived win rates obtained from Chatbot Arena, which reflect human preferences. 

\newparagraph{Question 3.} Does \our boost model preference tuning, resulting in a better alignment of the post-trained model?

$\bullet$ We train a DPO model using the \our calibrated reward model. By analyzing the downstream performance of the resulting policy model, we directly evaluate the effectiveness of \our in downstream post-training alignment.

\subsection{Implementation Details} \label{PreferenceDatasetConstruction}

For preference dataset construction, we use Preference700K \citep{dong2024rlhf}, a comprehensive dataset that aggregates preference data from eight sources. We randomly sampled 20K instructions from Preference700K and generated corresponding responses using selected over-valued and reference models. These responses were then scored by a reward model that we later calibrate. This process produced the uncalibrated preference dataset, which served as the foundation for applying \our to construct the calibrated preference dataset.

We set temperature $\tau$ = 0.7 and Top\_p = 0.9 during inference. We selected five reward models for calibration: \textit{Skywork-Reward-Llama-3.1-8B-v0.2} \citep{liu2024skywork}, \textit{RM-Mistral-7B}, \textit{FsfairX-LLaMA3-RM-v0.1} \citep{dong2023raft,xiong2024iterative}, \textit{GRM-llama3-8B-distill} \citep{yang2024regularizing}, and \textit{BTRM-Qwen2-7b-0613}. During reward model fine-tuning, we used the Adam optimizer \citep{kingma2014adam} with a learning rate of 2e-6, a weight decay of 0.001, and a cosine learning rate scheduler. The models were trained for 1 epoch.

\subsection{Experiment Results}

\subsubsection{Results on Reward Model Benchmarks}
\begin{table*}[t]
\small
\centering
\caption{Results of different reward models on the RM-Bench and RewardBench benchmark. We trained each RM on uncalibrated or CHARM-calibrated dataset. {$\uparrow(\downarrow)$} indicates the improvement (degradation) compared to original reward model. \our enhances overall performance, especially in Chat domain.}
\setlength{\tabcolsep}{0.85ex}
\begin{tabular}{@{}lclllllllll@{}}
\toprule
\multicolumn{1}{c}{\multirow{2}{*}{\textbf{\textsc{Reward Models}}}}    & \multirow{2}{*}{\textbf{\begin{tabular}[c]{@{}c@{}}\textsc{Mismatch}\\ \textsc{Degree}\end{tabular}}} & \multicolumn{8}{c}{\textbf{\textsc{RM-Bench}}}                                                                                                                                                                                                                                                                                                   & \multicolumn{1}{c}{\textbf{\textsc{RewardBench}}}                                         \\
\multicolumn{1}{c}{}                                           &                                                                                     & \multicolumn{1}{c}{Chat}                                                       & \multicolumn{1}{c}{Math} & \multicolumn{1}{c}{Code} & \multicolumn{1}{c}{Safety} & \multicolumn{1}{c}{Hard} & \multicolumn{1}{c}{Normal} & \multicolumn{1}{c}{Easy} & \multicolumn{1}{c}{\textit{Avg}}                                                 & \multicolumn{1}{c}{Chat-Hard}  \\ \midrule
\textsc{Skywork-RM} & \multirow{3}{*}{0.639} & 68.7 & 62.0 & 52.8 & 95.9 & 47.5 & 73.7 & 88.4 & 69.9 & 88.8 \\
\quad \textit{w/o calibration} &  & 68.9 & 61.9 & 53.1 & 95.9 & 47.6 & 73.8 & 88.5 & 70.0 & 88.8 \\
\quad \textit{w/ calibration} &  & \textbf{73.9}\textsubscript{(5.2$\uparrow$)} & 62.4 & 53.9 & 95.8 & 49.3 & 75.8 & 89.4 & \textbf{71.5}\textsubscript{(1.6$\uparrow$)} & \textbf{89.4}\textsubscript{(0.6$\uparrow$)} \\ \cmidrule(r){1-2}
\textsc{FsfairX-RM} & \multirow{3}{*}{0.554} & 62.5 & 63.2 & 54.6 & 90.4 & 44.9 & 71.6 & 86.5 & 67.7 & 65.3 \\
\quad \textit{w/o calibration} &  & 63.1 & 63.4 & 53.6 & 90.4 & 45.8 & 71.6 & 85.4 & 67.6 & 65.1 \\
\quad \textit{w/ calibration} &  & \textbf{64.5}\textsubscript{(2$\uparrow$)} & 63.3 & 56.0 & 90.0 & 45.4 & 72.6 & 87.5 & \textbf{68.5}\textsubscript{(0.8$\uparrow$)} & \textbf{65.7}\textsubscript{(0.4$\uparrow$)} \\ \cmidrule(r){1-2}
\textsc{Mistral-RM} & \multirow{3}{*}{0.528} & 60.8 & 56.6 & 52.6 & 88.7 & 37.5 & 68.2 & 88.3 & 64.7 & 60.5 \\
\quad \textit{w/o calibration} &  & 61.4 & 57.4 & 53.0 & 88.9 & 40.0 & 68.8 & 86.7 & 65.2 & 62.5 \\
\quad \textit{w/ calibration} &  & \textbf{63.2}\textsubscript{(2.4$\uparrow$)} & 57.0 & 52.4 & 88.3 & 36.3 & 69.8 & 89.5 & \textbf{65.2}\textsubscript{(0.5$\uparrow$)} & \textbf{65.1}\textsubscript{(4.6$\uparrow$)} \\ \cmidrule(r){1-2}
\textsc{GRM-RM} & \multirow{3}{*}{0.508} & 63.6 & 62.0 & 56.9 & 89.1 & 49.6 & 71.8 & 82.2 & 67.9 & 68.4 \\
\quad \textit{w/o calibration} &  & 63.6 & 62.4 & 58.3 & 89.5 & 49.8 & 72.7 & 82.9 & 68.4 & 68.8 \\
\quad \textit{w/ calibration} &  & \textbf{66.2}\textsubscript{(2.6$\uparrow$)} & 62.6 & 58.0 & 89.3 & 48.3 & 73.9 & 84.9 & \textbf{69.0}\textsubscript{(1.1$\uparrow$)} & \textbf{68.9}\textsubscript{(0.5$\uparrow$)} \\ \cmidrule(r){1-2}
\textsc{BTRM-RM} & \multirow{3}{*}{0.162} & 60.0 & 61.3 & 53.8 & 89.9 & 37.1 & 71.0 & 90.7 & 66.3 & 58.1 \\
\quad \textit{w/o calibration} &  & 58.5 & 61.5 & 54.1 & 89.1 & 35.3 & 70.8 & 91.3 & 65.8 & 58.7 \\
\quad \textit{w/ calibration} &  & \textbf{60.2}\textsubscript{(0.2$\uparrow$)} & 60.5 & 53.8 & 89.6 & 34.8 & 71.1 & 92.1 & \textbf{66.0}\textsubscript{(0.3$\downarrow$)} & \textbf{57.8}\textsubscript{(0.3$\downarrow$)} \\ \bottomrule
\end{tabular}
\label{RewardModelBenchmark}
\vspace{-2mm}
\end{table*}


After obtaining reward scores for various policy models on AlpacaEval, as described in Section~\ref{PreferenceBiasinRewardModels}, we derive a scoring profile for each model. We then select a strong model \textit{GPT-4o-mini} as our reference and compute the Mismatch Degree of every other policy model relative to it. Finally, we choose the model pair with the highest Mismatch Degree, \textit{gemma-2-9b-it-SimPO} vs. \textit{GPT-4o-mini}, for our further calibration.

We choose five reward models from the RM-Bench leaderboard, each exhibiting varying levels of performance. We compute their MD on the selected model pair, revealing distinct deviations in how they value \textit{Gemma-2-9b-it-SimPO} relative to human preferences. These reward models serve as the base models for our experiments. Following the methodology described in Section~\ref{PreferenceDatasetConstruction}, we construct both uncalibrated and calibrated preference datasets for each reward model.

We evaluated three versions of each reward model on the benchmark: \textbf{(1)} the original reward model, \textbf{(2)} the reward model trained on the uncalibrated dataset, and \textbf{(3)} the reward model trained on the calibrated dataset. We select RM-Bench and RewardBench as our test benchmarks. For RewardBench, only the more challenging Chat-Hard domain is reported since other domains have shown nearly saturated results. The overall results are displayed in Table~\ref{RewardModelBenchmark}.

From the benchmark results across different versions of the reward model, we can summarize the following findings:

\newparagraph{Finding 1: Uncalibrated Preference Datasets Lead to Minimal Performance Gains.} Across all evaluated models, uncalibrated training led to minimal or no improvement over the original reward model. While an uncalibrated preference dataset introduces additional preference data, it does not explicitly correct biases. The underlying issues in the RM's decision boundaries remain unaddressed, resulting in no meaningful shift in performance.

\newparagraph{Finding 2: \our Enhances Overall Performance, Especially in Chat Domain.} Training on the calibrated preference dataset enhances reward model performance across benchmarks. On average, RM-Bench scores improved by +0.74 points, with Skywork-RM showing the largest gain of +1.6 points. Among all evaluated tasks, Chat performance saw the most substantial improvement after calibration. Skywork-RM achieved the largest gain of +5.2 points, followed by Mistral-RM of +2.4 points and FsfairX-RM of +2.0 points. And a similar trend was observed in RewardBench Chat-Hard. 

\subsubsection{Results on Human Preference Alignment}

\begin{figure*}[t]
    \centering
    \includegraphics[width=0.85\linewidth]{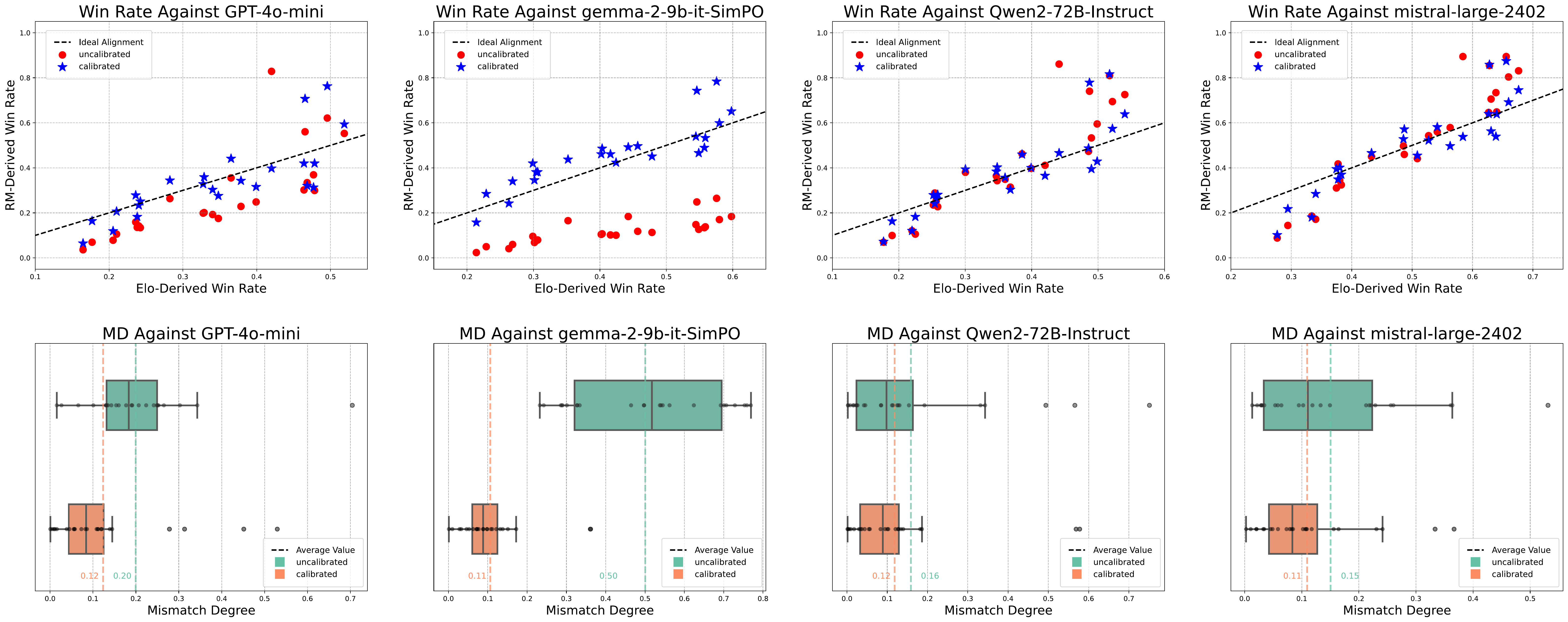}
    \caption{Win rates and Mismatch Degrees before and after calibration. In the win rate plots, the x-axis is the expected win rates calculated based on the models' Elo scores, while the y-axis is the win rates derived from the reward model scores. Points closer to the dotted line indicate a better alignment between the reward model and human preferences.}
    \label{Calibrate}
    \vspace{-2mm}
\end{figure*}

One of the primary objectives of our calibration method is to better align the reward model's judgment with human preferences. To evaluate whether \our effectively mitigates model preference bias, we designed experiments on the AlpacaEval dataset, which contains a wide range of policy models and their responses, making it convenient for constructing pairwise battles. We selected 24 policy models and their responses from AlpacaEval and then used both the original and calibrated Skywork-RM to score these responses. After scoring, we built pairwise comparisons from the RM-assigned scores to obtain RM-derived win rates between model pairs. These win rates were compared against Elo-derived win rates, allowing us to compute the Mismatch Degree as a measure of model preference bias. The results are presented in Figure~\ref{Calibrate}.

\begin{table*}[]
\small

\caption{Performance of Qwen2.5-7B-Instruct on IFEval under different DPO settings. We use original Skywork-RM and CHARM-calibrated Skywork-RM to produce DPO datasets. {$\uparrow(\downarrow)$} indicates the improvement (degradation) compared to Qwen2.5-7B-Instruct.}
\centering
\begin{tabular}{@{}lllll@{}}
\toprule
\multicolumn{1}{c}{\multirow{2}{*}{\textbf{\textsc{Models}}}} & \multicolumn{4}{c}{\textbf{\textsc{IFEval}}}                       \\
\multicolumn{1}{c}{}                                 & Inst. Loose & Inst. Strict & Prompt Loose & Prompt Strict \\ \midrule
\textsc{Qwen2.5-7B-Instruct}                                  & 72.06       & 67.51        & 62.11        & 56.75         \\
\quad \textit{w/ SkyworkRM DPO}                                    & 72.30       & 67.87        & 62.29        & 56.19         \\
\quad \textit{w/ CHARM DPO}                           & \textbf{74.46}\textsubscript{(2.4$\uparrow$)}       & \textbf{68.47}\textsubscript{(0.96$\uparrow$)}        & \textbf{64.88}\textsubscript{(2.77$\uparrow$)}        & \textbf{57.67}\textsubscript{(0.92$\uparrow$)}         \\ \bottomrule
\end{tabular}
\label{dpo}
\vspace{-2mm}
\end{table*}

\newparagraph{Finding 3: \our Reduces Model Preference Bias.} From the win rate comparison, we observe that models' performance against \textit{GPT-4o-mini-2024-07-18} and \textit{Gemma-2-9b-it-SimPO} exhibits stronger alignment with ideal human preferences. Specifically, the win rates derived from RM scores are now closer to those based on Elo scores. Additionally, we compute the MD across different models, and the results reveal a clear reduction in MD. This indicates that calibration effectively mitigates model preference bias for both the over-valued and reference models, demonstrating improved alignment of calibrated reward models with human preferences.

\newparagraph{Finding 4: \our Generalizes to Unseen Models.} We further selected \textit{Qwen2-72B-Instruct} and \textit{mistral-large-2402}, two policy models that were not used during the calibration process, to assess whether our method generalizes to unseen LLMs. Results in Figure~\ref{Calibrate} (right) show that the calibrated reward model maintains a stronger correlation with human preferences even on these unseen models. Notably, this generalization emerges despite using only two policy models during calibration, suggesting the existence of more inherent factors underlying model preference bias. We make further analysis in the following section.

\subsubsection{Results on Post Training Alignment}

To determine whether \our offers practical post-training alignment benefits beyond improvements on RM benchmarks, we evaluate its impact in a real post-training setting. Specifically, we trained a DPO model using the \our-calibrated reward model. Since model preference bias may originate from biased preference-learning datasets as discussed in Appendix~\ref{appendix_preference_bias}, offline DPO provides an ideal controlled environment for assessing the effectiveness of \our.

We generated responses from \textit{GPT-4o-mini} and \textit{Gemma-2-9b-it-SimPO} on Preference20K and obtained reward scores from both the original Skywork-RM and the \our-calibrated Skywork-RM. Using these scores, we constructed two DPO datasets and trained \textit{Qwen2.5-7B-Instruct} separately on each. The resulting models were evaluated on IFEval, and the results are shown in Table~\ref{dpo}.

\newparagraph{Finding 5: \our Improves Downstream Model Performance.} Models trained with \our-calibrated RM consistently outperform those trained with the original RM and the baseline model across all IFEval metrics. Notably, the uncalibrated RM provides marginal gains or even degrades performance, while \our yields consistent improvements. These results demonstrate that correcting model preference bias is beneficial for enhancing downstream model performance.

\section{Analysis}\label{analysis}
\begin{table*}[]\small
\centering
\caption{Impact of Mismatch Degree on calibration effectiveness. We construct different model pairs with varing MD and perform \our on them. \textbf{Bold} indicates the best results, \underline{Underline} denotes the second-best results.}
\begin{tabular}{@{}cccccccc@{}}
\toprule
\multirow{2}{*}{\textbf{\textsc{Over-valued Models}}} & \multirow{2}{*}{\textbf{\textsc{Ref Models}}}       & \multirow{2}{*}{\textbf{\begin{tabular}[c]{@{}c@{}}\textsc{Mismatch}\\ \textsc{Degree}\end{tabular}}} & \multicolumn{4}{c}{\textbf{\textsc{RM-Bench}}}                         & \multirow{2}{*}{\textbf{\textsc{Avg.}}} \\
                                             &                                            &                                                                                     & Chat          & Math          & Code          & Safety        &                               \\ \midrule
\multicolumn{2}{c}{\textbf{\textsc{Skywork-RM}}}                                                   &                                                                                     & 68.7          & 62.0          & 52.8          & 95.9          & 69.9                          \\ \midrule
\multicolumn{1}{l}{\textsc{\textsc{Gemma-2-9b-it}-SimPO}}      & \multirow{6}{*}{\textsc{GPT-4o-mini}}               & 0.639                                                                               & \textbf{73.9} & 62.4          & 53.9          & 95.8          & \underline{71.5}                          \\
\multicolumn{1}{l}{\textsc{Gemma-2-27b-it}}           &                                            & 0.225                                                                               & 70.7          & 61.9          & 52.9          & \textbf{96.7} & 70.5                          \\
\multicolumn{1}{l}{\textsc{Gemma-2-9b-it}}            &                                            & 0.155                                                                               & 70.9          & 62.2          & 52.9          & \underline{96.6}          & 70.7                          \\
\multicolumn{1}{l}{\textsc{Qwen2.5-72B-Instruct}}     &                                            & 0.088                                                                               & 70.5          & 62.1          & 52.6          & 96.2          & 70.4                          \\
\multicolumn{1}{l}{\textsc{Llama-3.1-70B-Instruct}}   &                                            & 0.048                                                                               & 68.9          & 61.9          & 53.0          & 96.0          & 70.0                          \\
\multicolumn{1}{l}{\textsc{Llama-3.1-8B-Instruct}}    &                                            & 0.032                                                                               & 68.8          & 61.6          & 52.5          & 96.2          & 69.8                          \\ \midrule
\multirow{5}{*}{\textsc{\textsc{Gemma-2-9b-it}-SimPO}}         & \multicolumn{1}{l}{\textsc{Gemma-2-9b-it}}          & 0.582                                                                               & 70.7          & 63.2          & \underline{54.6}          & 94.7          & 70.8                          \\
                                             & \multicolumn{1}{l}{\textsc{Gemma-2-27b-it}}         & 0.633                                                                               & 70.0          & 63.1          & 54.0          & 94.7          & 70.4                          \\
                                             & \multicolumn{1}{l}{\textsc{Llama-3.1-8B-Instruct}}  & 0.506                                                                               & 71.4          & \underline{64.5}          & 53.7          & 95.9          & 71.4                          \\
                                             & \multicolumn{1}{l}{\textsc{Llama-3.1-70B-Instruct}} & 0.675                                                                               & \underline{72.2}          & \textbf{64.7} & \textbf{55.0} & 95.2          & \textbf{71.8}                 \\
                                             & \multicolumn{1}{l}{\textsc{Qwen2.5-72B-Instruct}}   & 0.625                                                                               & 70.9          & 62.4          & 53.8          & 96.0          & 70.8                          \\ \bottomrule
\end{tabular}
\label{MismatchDegreeExtend}
\vspace{-2mm}
\end{table*}
\subsection{Mismatch Degree Serves as an Indicator of Calibration Need}

We observe a potential correlation between Mismatch Degree and performance improvement after calibration in Table~\ref{RewardModelBenchmark}. Notably, Skywork-RM, which exhibited the highest MD, achieved the most significant performance gains. In contrast, BTRM, which had the lowest MD, even experienced a slight performance degradation. To further investigate the relationship between MD and calibration effectiveness, we design additional experiments to analyze how MD influences the impact of reward model calibration. We fix Skywork-RM as the reward model and construct different model pairs to produce varying levels of mismatch degree. We then repeat the same experiments on each model pair. The results are presented in Table~\ref{MismatchDegreeExtend}.

By analyzing the results, we observe that MD serves as a strong indicator of a model’s misalignment and the potential benefits of calibration. Models with higher MD values tend to exhibit greater improvements after calibration. For instance, \textit{gemma-2-9b-it-SimPO} vs. \textit{GPT-4o-mini} (MD = 0.639) and \textit{gemma-2-9b-it-SimPO} vs. \textit{Llama-3.1-70B-Instruct} (MD = 0.675) benefit the most from calibration, showing significant performance gains. Conversely, models with near-zero MD, such as \textit{Qwen2.5-72B-Instruct} vs. \textit{GPT-4o-mini} (MD = 0.088) and \textit{Llama-3.1-8B-Instruct} vs. \textit{GPT-4o-mini} (MD = 0.032), experience minimal or even negative performance changes after calibration. This finding highlights that if a model is already well-aligned with human preferences, additional calibration may have little effect or even introduce instability. To precisely quantify this relationship, we calculated the Pearson correlation between Mismatch Degree and its average performance improvement. The result is a high correlation of 0.747, confirming that MD is a practical tool for diagnosing mis-calibration in reward models.

\subsection{Category-wise CHARM with More Nuanced Correction}\label{categorywise}

\begin{table}[]\small
\centering
\caption{Mismatch Degree for each category.}
\begin{tabular}{@{}lc@{}}
\toprule
\multicolumn{1}{c}{\multirow{2}{*}{\textbf{\textsc{Category}}}} & \multirow{2}{*}{\textbf{\begin{tabular}[c]{@{}c@{}}\textsc{Mismatch}\\ \textsc{Degree}\end{tabular}}} \\
\multicolumn{1}{c}{}                                   &                                                                                     \\ \midrule
Code Generation                                        & 0.736                                                                               \\
Creative Writing                                       & 0.538                                                                               \\
Factual QA                                             & 0.894                                                                               \\
Instruction Following                                  & 0.411                                                                               \\
Math/Reasoning                                         & 0.322                                                                               \\
Others                                                 & 0.875                                                                               \\ \bottomrule
\end{tabular}
\label{appendix_category_md}
\vspace{-6mm}
\end{table}

Model preference bias operates at the distributional level, and overall scoring behavior can be viewed as a mixture of category-specific distributions. While fine-grained, category-wise calibration is a natural extension, it introduces additional complexity in labeling and processing categories.

To explore this, we conducted an experiment using GPT-5 to categorize prompts in our calibration dataset into six types: \textbf{Math/Reasoning}, \textbf{Creative Writing}, \textbf{Code Generation}, \textbf{Factual QA}, \textbf{Instruction Following}, and \textbf{Others}. For each category, we computed the Mismatch Degree and optimized the offset between \textit{Gemma-2-9b-it-SimPO} and \textit{GPT-4o-mini} using Skywork-RM. \our was then applied separately within each category, and the calibrated datasets were merged for reward-model training. Results see Table~\ref{appendix_category_md} and Table~\ref{appendix_category_rmbench}.

Category-wise calibration yields an additional improvement of 0.9 points on RM-Bench, which is consistent with expectations. This indicates that for scenarios where maximizing performance is the primary goal, category-wise calibration can provide further gains, and \our naturally extends to this setting. However, as our main contribution emphasizes a simple and efficient calibration method, we adopt a single global offset in the main paper, highlighting the tradeoff between simplicity and fine-grained calibration.

\subsection{Enhanced Robustness to Stylistic Variations}
We observe that \our leads to consistent performance improvements on RM-Bench. This benchmark is particularly challenging because the differences between preferred and rejected responses are often subtle, and the responses exhibit diverse stylistic variations ranging from concise and straightforward to elaborate and well-formatted. Achieving higher scores on RM-Bench therefore suggests that the reward model becomes more robust to stylistic changes and better focuses on the factual reliability and coherence of responses rather than being misled by superficial patterns.

Based on this observation, we hypothesize that \our encourages the RM to prioritize semantic correctness and content reliability over superficial formatting that is frequently associated with certain policy models \citep{zhang2024lists}. We conducted a analysis of the RM’s sensitivity to stylistic variation. We first applied Z-score normalization across all RM-Bench samples and then computed the average normalized scores that the RM assigned to the responses across three distinct style categories. The results show that after calibration, the variance of average Z-scores across chosen and rejected groups decreased from 0.123 to 0.099 and 0.058 to 0.049, reflecting \our  implicitly enhances the robustness of reward models to stylistic variations.

\begin{table}[]\small
\caption{RM-Bench performance of different offset type calibration. {$\uparrow(\downarrow)$} indicates the improvement (degradation) compared to Skywork-RM.}
\centering
\resizebox{\linewidth}{!}{%
\begin{tabular}{@{}llllll@{}}
\toprule
\multicolumn{1}{c}{\multirow{2}{*}{\textbf{\textsc{Offset Type}}}} & \multicolumn{5}{c}{\textbf{\textsc{RM-Bench}}}      \\
\multicolumn{1}{c}{}                                      & Chat & Math & Code & Safety & \textit{Avg} \\ \midrule
\textsc{Skywork-RM}                                                & 68.7 & 62.0 & 52.8 & 95.9   & 69.9         \\
\quad \textit{w/ global-wise}                                      & 72.5 & 62.9 & 54.4 & 95.7   & 71.4         \\
\quad \textit{w/ category-wise}                                    & 74.2\textsubscript{(5.5$\uparrow$)} & 64.2\textsubscript{(2.2$\uparrow$)} & 54.5\textsubscript{(1.7$\uparrow$)} & 96.2\textsubscript{(0.3$\uparrow$)}   & 72.3\textsubscript{(2.4$\uparrow$)}         \\ \bottomrule
\end{tabular}
}
\label{appendix_category_rmbench}
\vspace{-4mm}
\end{table}

\subsection{From Explicit Patterns to Implicit Calibration}
Based on our previous analysis, we conducted a further quantitative study on five specific stylistic patterns: emoji, length, bold, exclamation, and list. We calculated the frequency of these patterns appearing in chosen and rejected responses within the dataset we used. We introduce the Preference Ratio, defined as the ratio of a pattern's occurrences in chosen responses to its occurrences in rejected responses.

The analysis reveals that the most significant changes occurred in two strong stylistic features: emojis and bold formatting. Before calibration, the reward model exhibited a strong positive bias: its preference ratio for responses containing emojis was as high as 2.38, and for those with bold text, it was 1.25. After calibration with \our, this preference underwent a reversal. 


This result reflects that \our can be understood mechanistically as a form of implicit calibration. The core point is that a reward model's preference for a specific policy model is, in essence, an inherited preference for patterns present in that model's outputs. However, these patterns extend far beyond the few explicit features we have listed; they more likely include a large number of subtle, implicit linguistic styles that are difficult to identify and enumerate through manual construction. By calibrating the overall score distribution, we can systematically mitigate the reward model's over-reliance on these complex patterns without needing to explicitly identify and construct each biased pattern. 

\begin{table}[t]\footnotesize
\centering
\caption{Statistics of stylistic pattern occurrences in the calibration dataset. CH. refers to chosen, RE. refers to rejected and PR. refers to the ratio of a pattern’s occurrences in chosen responses to its occurrences in rejected
responses.}
\resizebox{\linewidth}{!}{%
\begin{tabular}{lcccccc}
\toprule
\multirow{2}{*}{\textbf{\textsc{Pattern}}} & \textbf{\textsc{CH.}} & \textbf{\textsc{RE.}} & \textbf{\textsc{PR.}} & \textbf{\textsc{CH.}} & \textbf{\textsc{RE.}} & \textbf{\textsc{PR.}} \\
 & \multicolumn{3}{c}{\textit{w/o calibration}} & \multicolumn{3}{c}{\textit{w/ calibration}} \\ \midrule
Emoji & 357 & 150 & 2.38 & 193 & 313 & 0.61 \\
Length & 1842 & 1752 & 1.05 & 1851 & 1742 & 1.06 \\
Bold & 16595 & 13313 & 1.24 & 13961 & 15941 & 0.87 \\
Excl. & 5541 & 4705 & 1.17 & 5609 & 4635 & 1.21 \\
List & 10546 & 12457 & 0.84 & 11438 & 11565 & 0.98 \\ \bottomrule
\end{tabular}
}
\label{dataset_stat}
\vspace{-4mm}
\end{table}

\section{Discussions}




While our work focuses on discriminative reward models, we acknowledge the existence of alternative formulations, such as generative \citep{zhang2024genrm} reward modeling. To explore the pervasiveness of this bias, we conducted a preliminary probe into generative reward models (see Appendix~\ref{appendix_grm}). Our findings indicate that model preference bias is also present in GRMs. This suggests that the bias is not confined to scalar reward modeling; even generative RMs, often considered more interpretable, are susceptible to exploiting spurious implicit patterns. Reasoning models enhanced with reinforcement learning \citep{qwen3technicalreport,guo2025deepseek} have demonstrated greater robustness and less influenced model preference bias, suggesting a promising direction for building more reliable reward models. 



\section{Conclusion}


In this work, we identified a subtle yet consequential bias in widely used reward models, which we term model preference bias. This bias reflects a tendency for RMs to disproportionately overvalue responses from certain policy models, resulting in misaligned evaluations and unfair judgments. To address this issue, we introduced \our, a simple yet effective calibration method that leverages Elo scores from Chatbot Arena to construct debiased preference datasets.

Extensive experiments demonstrate that \our enhances the judging capability of reward models, yielding consistent performance gains on benchmarks such as RM-Bench and RewardBench, particularly in the chat domain. It also improves alignment with human preferences, mitigates model preference bias and improves downstream post-training performance. Our findings highlight the need to further explore the mechanisms underlying model preference bias towards developing more robust and reliable reward models.
\clearpage

\section*{Impact Statement}

This paper presents work whose goal is to advance the field of Machine
Learning. There are many potential societal consequences of our work, none
which we feel must be specifically highlighted here.
\bibliography{example_paper}
\bibliographystyle{icml2026}

\newpage
\appendix
\appendix
\section{Appendix}
\subsection{More Experiments on Model Preference Bias}\label{appendix_preference_bias}

\begin{figure}[ht]
    \centering
    \includegraphics[width=0.95\linewidth]{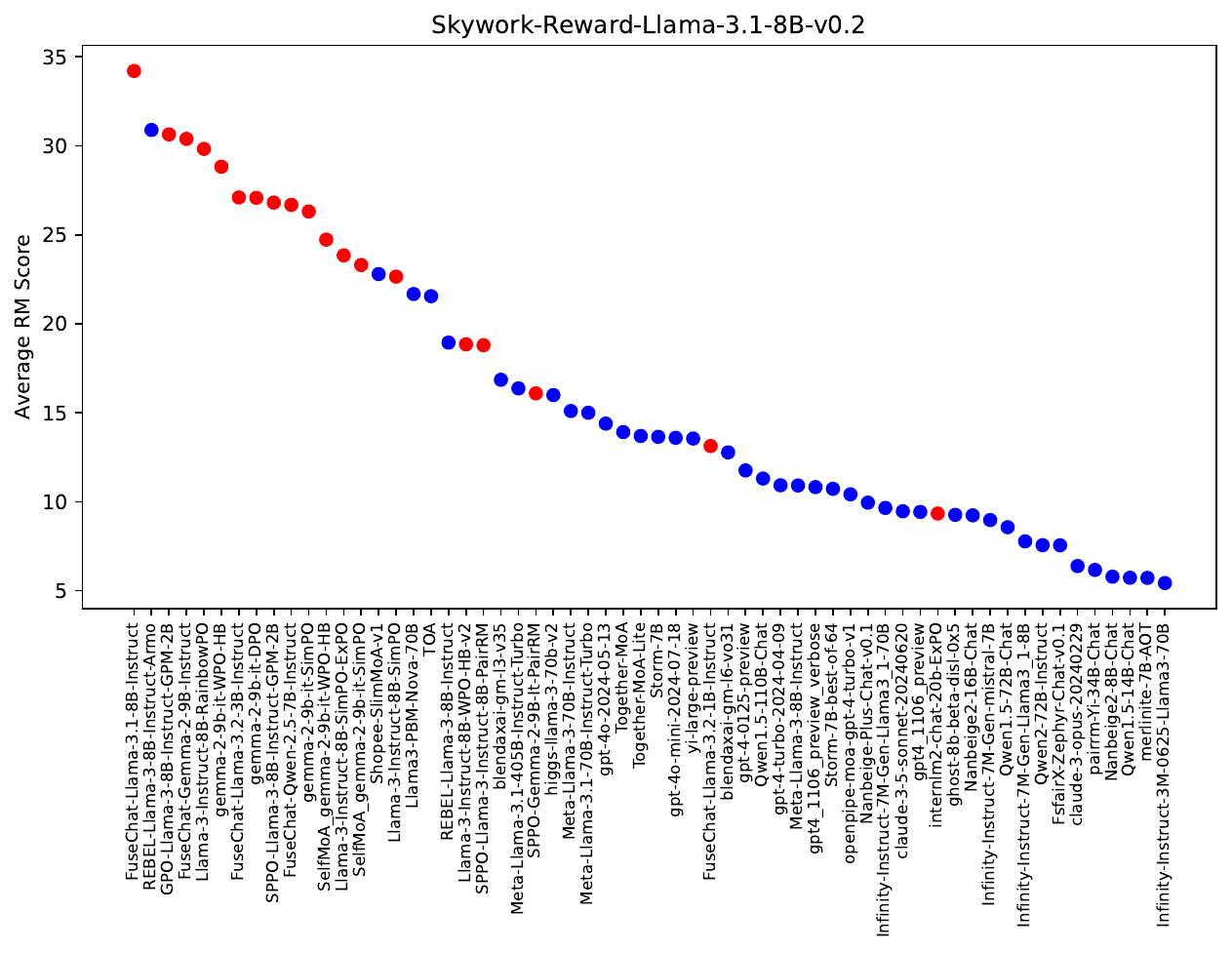} 
    \caption{Score results of Skywork-RM on more policy models in the AlpacaEval dataset.}
    \label{rm_skywork}
\end{figure}

\begin{figure}[ht]
    \centering
    \includegraphics[width=0.95\linewidth]{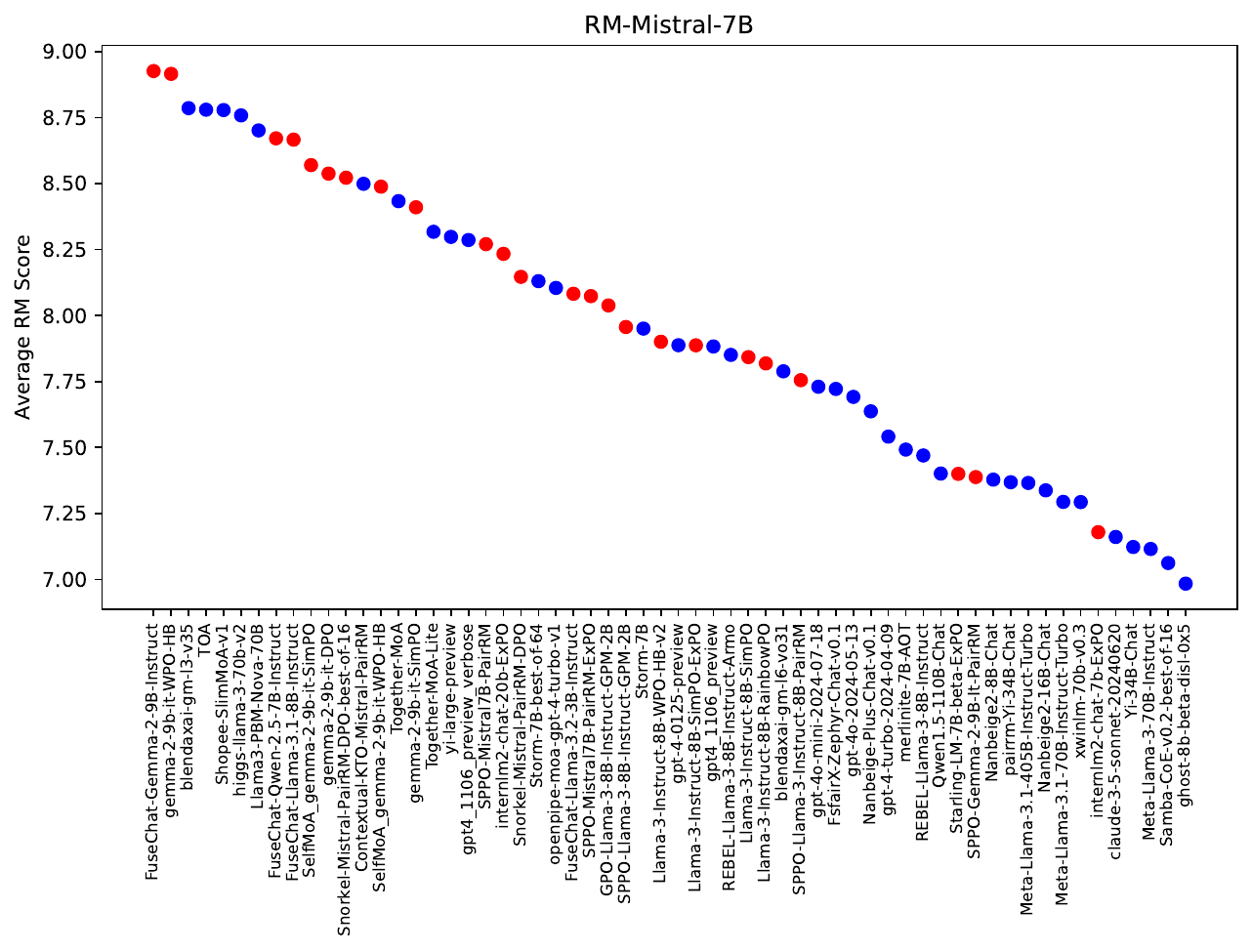} 
    \caption{Score results of Mistral-RM on more policy models in the AlpacaEval dataset.}
\end{figure}

\begin{figure}[ht]
    \centering
    \includegraphics[width=0.95\linewidth]{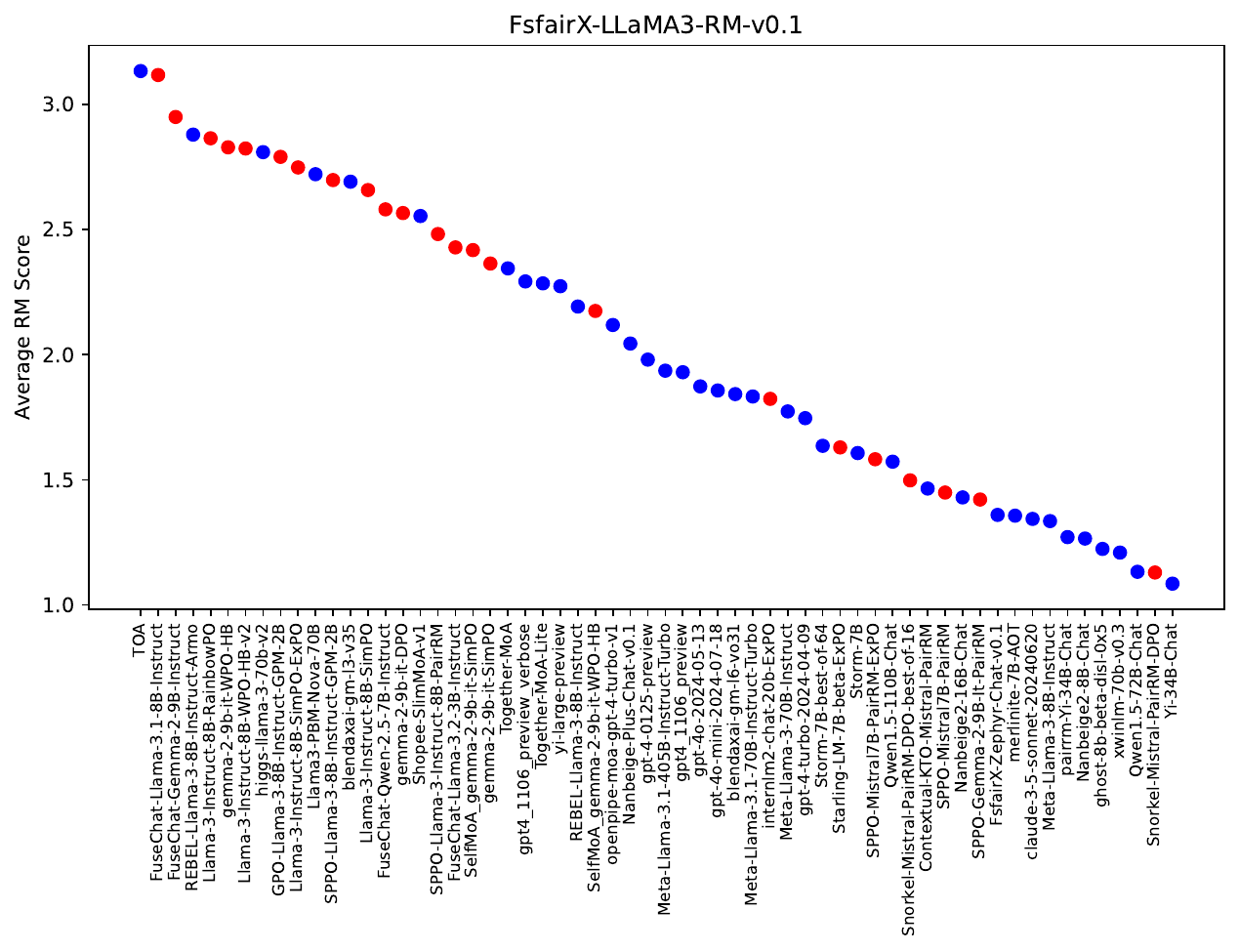}
    \caption{Score results of FsfairX-RM on more policy models in the AlpacaEval dataset.}
\end{figure}

\begin{figure}[ht]
    \centering
    \includegraphics[width=0.95\linewidth]{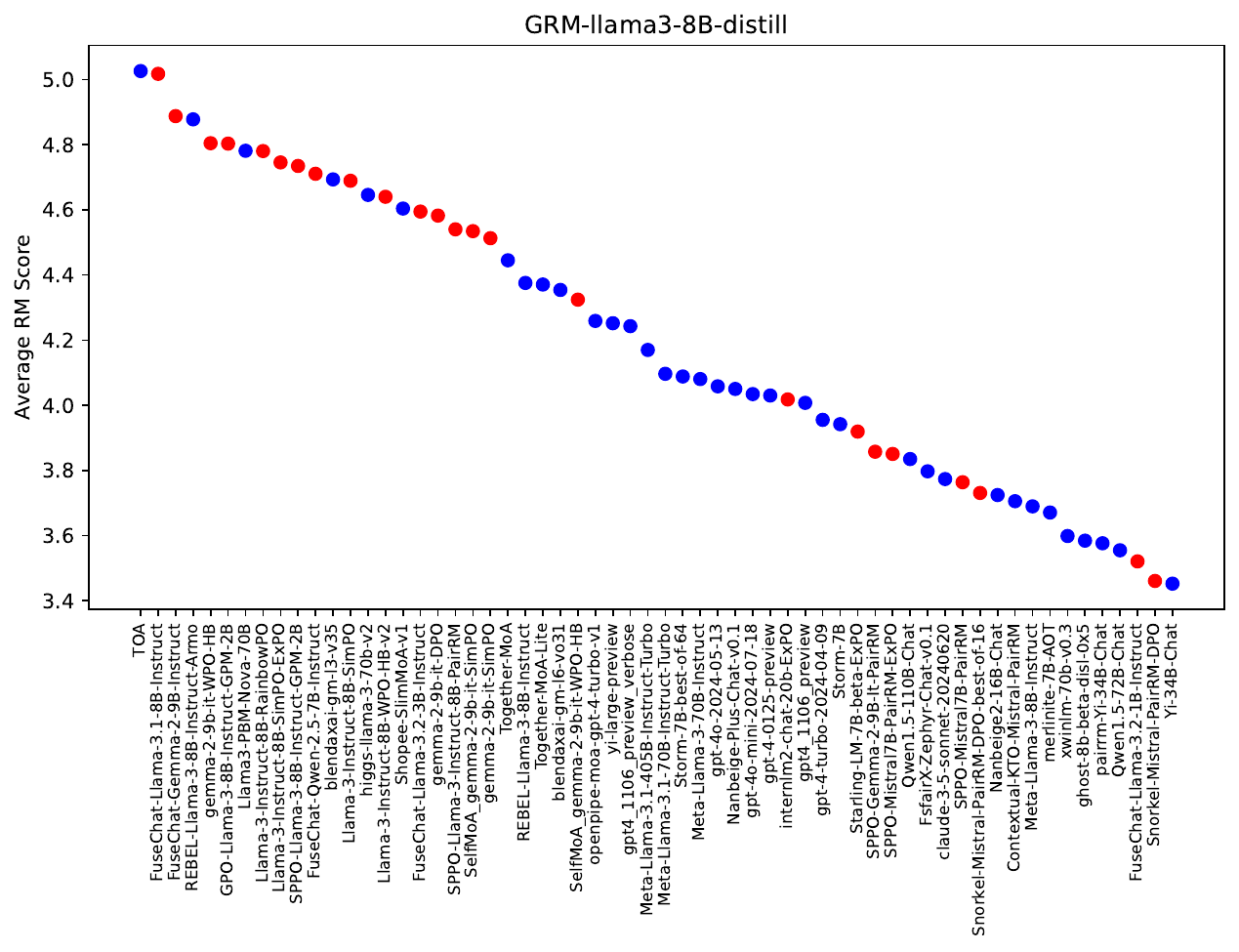} 
    \caption{Score results of GRM-RM on more policy models in the AlpacaEval dataset.}
\end{figure}

\begin{figure}[t]
    \centering
    \includegraphics[width=0.95\linewidth]{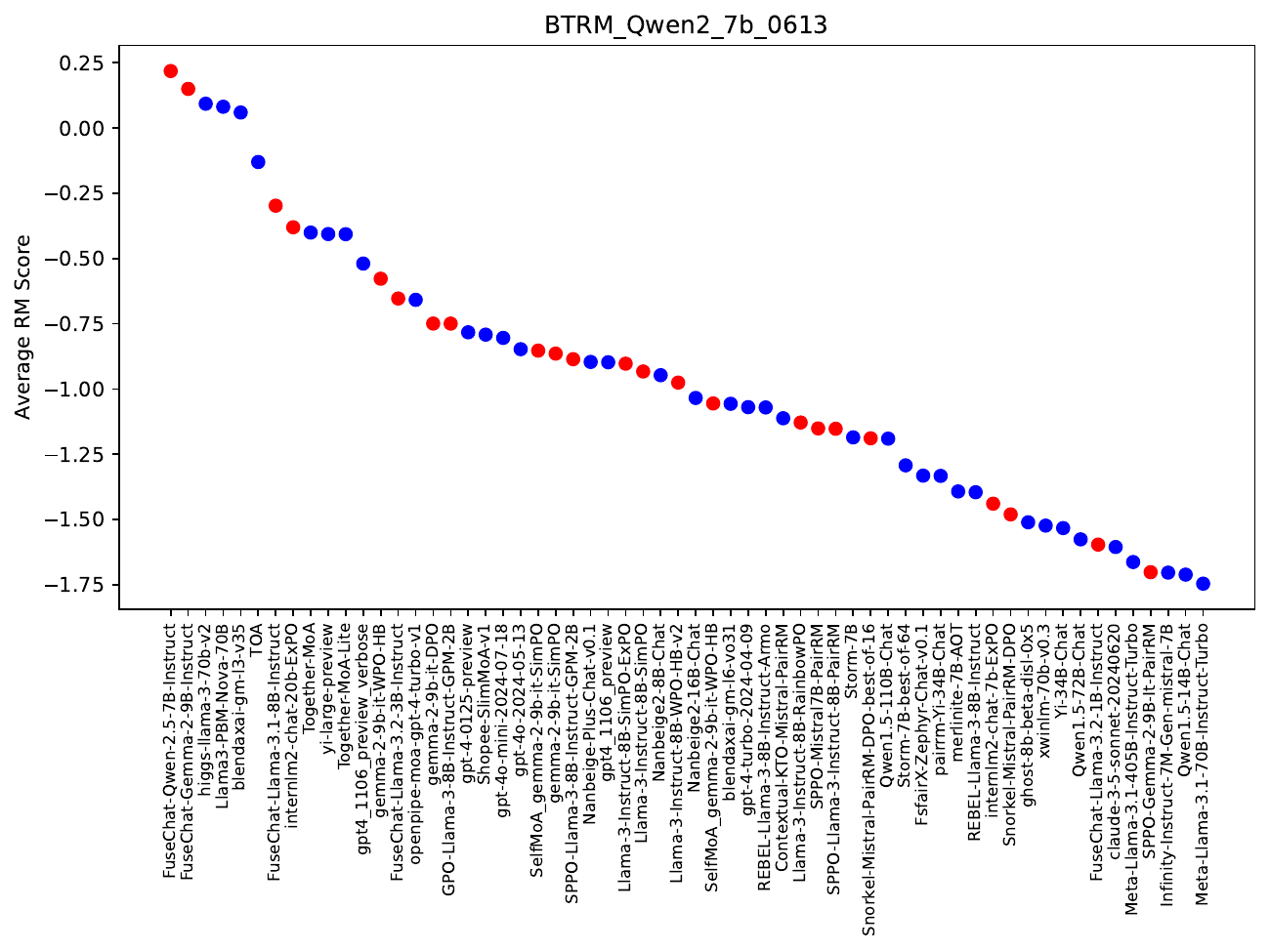} 
    \caption{Score results of BTRM-RM on more policy models in the AlpacaEval dataset.}
    \label{rm_btrm}
\end{figure}

\clearpage

In Section~\ref{PreferenceBiasinRewardModels}, we compare the correlation between RM scores assigned to policy models and their Arena Elo scores. However, since many models listed on the AlpacaEval leaderboard have not participated in Chatbot Arena, their Elo scores are unavailable, preventing direct comparison with human preferences.

To further analyze model preference bias, we score responses from all 228 models available in the AlpacaEval dataset using different RMs. We then select the top 60 models ranked by Average RM scores. The results are illustrated in Figures~\ref{rm_skywork}--\ref{rm_btrm}, where models marked in red indicate those that have undergone preference optimization. These models are mostly around 7B parameters, significantly smaller than the top-ranking commercial models in Chatbot Arena. However, under RM evaluation, they exhibit a totally different ranking trend. 

This finding suggests that the model preference bias exhibited by RMs is not a preference for specific individual model, but rather a systematic preference for a class of models. This, in turn, points to the formation mechanism of this bias being linked to common training methodologies or shared datasets utilized by this particular class of models. We hope this observation provides a valuable direction for future research into the origins of such biases.

\subsection{Policy Model Details}\label{appendix_elo_stat}

\begin{table*}[]
\centering
\caption{Policy models used in our empirical study, along with their corresponding Chatbot Arena Elo scores and AlpacaEval win rates.}
\begin{tabular}{@{}lcc@{}}
\toprule
\multicolumn{1}{c}{\textbf{Policy Models}} & \textbf{Chatbot Arena Elo} & \textbf{AlpacaEval Winrate} \\ \midrule
gpt-4o-2024-05-13 & 1285 & 57.5 \\
gpt-4o-mini-2024-07-18 & 1272 & 50.7 \\
Meta-Llama-3.1-405B-Instruct-Turbo & 1269 & 39.3 \\
Qwen2-72B-Instruct & 1257 & 38.1 \\
gpt-4-turbo-2024-04-09 & 1256 & 55 \\
gpt4\_1106\_preview & 1250 & 50 \\
Meta-Llama-3.1-70B-Instruct-Turbo & 1248 & 38.1 \\
claude-3-opus-20240229 & 1247 & 40.5 \\
gemma-2-9b-it-SimPO & 1216 & 72.4 \\
claude-3-sonnet-20240229 & 1201 & 34.9 \\
gpt4\_0314 & 1186 & 35.3 \\
Meta-Llama-3.1-8B-Instruct-Turbo & 1176 & 20.9 \\
gpt4\_0613 & 1163 & 30.2 \\
mistral-large-2402 & 1157 & 32.7 \\
mistral-medium & 1148 & 28.26 \\
Mixtral-8x22B-Instruct-v0.1 & 1147 & 30.9 \\
gemini-pro & 1110 & 24.4 \\
OpenHermes-2.5-Mistral-7B & 1074 & 16.2 \\
Qwen1.5-7B-Chat & 1070 & 14.7 \\
gpt-3.5-turbo-1106 & 1068 & 19.3 \\
vicuna-13b & 1042 & 9.2 \\
gemma-7b-it & 1037 & 10.4 \\
vicuna-7b & 1005 & 6.3 \\
gemma-2b-it & 990 & 5.4 \\ \bottomrule
\end{tabular}

\label{elo_stat}
\vspace{-2pt}
\end{table*}

In Section~\ref{PreferenceBiasinRewardModels}, we conducted an empirical study on a diverse set of policy models to probe model preference bias. This experiment relies on the Chatbot Arena Elo scores and the AlpacaEval win rates of the selected models. For completeness, we report the Elo scores and win rates used in our study in Table~\ref{elo_stat}. Since the Chatbot Arena leaderboard is dynamically updated, these values may differ slightly from the latest results, but such discrepancies do not affect the overall conclusions of our experiments.

\subsection{Model Preference Bias in GRMs}\label{appendix_grm}

\begin{table*}[]
\caption{Results of model preference bias on several GRMs.}
\centering
\begin{tabular}{@{}lccc@{}}
\toprule
\multicolumn{1}{c}{\textbf{GRMs}} & \textit{deepseek-v3} \textbf{wins} & \textit{gemma-2-9b-it-SimPO} \textbf{wins} & \textbf{MD} \\ \midrule
Llama-3.1-8B & 500 & 499 & 0.296 \\
DeepSeek-R1-Distill-Qwen-7B & 570 & 365 & 0.143 \\
Skywork-Critic-Llama-3.1-8B & 338 & 662 & 0.525 \\
JudgeLRM-7B & 563 & 432 & 0.204 \\
Qwen3-8B & 762 & 233 & 0.07 \\
DeepSeek-R1 & 783 & 213 & 0.104 \\ \bottomrule
\end{tabular}

\label{grm}
\end{table*}

Generative Reward Models (GRMs) are a type of reward model that directly utilizes the judging capabilities of LLMs. Through techniques like Chain-of-Thought prompting \citep{wei2022chain}, an LLM can be transformed into a vanilla GRM, also known as an "LLM-as-a-Judge". Furthermore, by fine-tuning on preference pairs, an LLM can be equipped with more robust reward modeling abilities \citep{zhu2023judgelm,chen2025judgelrm,liu2025inference}. GRMs are often considered more robust and interpretable than their scalar counterparts because the generative output format has a naturally higher information density at the cost of inference time.

This perceived robustness motivated our investigation into whether GRMs are also susceptible to the model preference bias identified in our work. To this end, we designed an experiment with three categories of models: (1) LLM-as-a-judge system utilizing CoT. (2) GRM fine-tuned on preference datasets. (3) Reasoning models enhanced with Reinforcement Learning.

For the experiment, we selected 1,000 prompts and let \textit{DeepSeek-V3} and \textit{Gemma-2-9B-it-SimPO} to generate responses. These response pairs were then judged by the different GRMs. To prevent positional bias, the order of the response pairs was shuffled before evaluation. The results are displayed in Table~\ref{grm}.

Both the LLM-as-a-judge system and the GRM fine-tuned on the preference dataset exhibited a discernible degree of model preference bias, with the \textit{Skywork-Critic-Llama-3.1-8B} model showing the most significant bias. In contrast, the two reasoning models performed excellently, achieving near-perfect alignment with human preferences. This suggests that while even generative models are vulnerable to model preference bias, advanced training methods like RL that enhance a model's reasoning capabilities may be a promising direction for mitigating such biases.

\subsection{Robustness to Chatbot Arena Elo Updates}

\begin{table*}[]
\caption{Robustness of CHARM to Elo rating updates. Despite substantial shifts in absolute Elo scores, calibration performance remains nearly identical.}
\centering
\begin{tabular}{@{}lccccccc@{}}
\toprule
\multicolumn{1}{c}{\multirow{2}{*}{\textbf{Elo Version}}} & \multirow{2}{*}{\textbf{Ref Elo}} & \multirow{2}{*}{\textbf{Over-valued Elo}} & \multicolumn{5}{c}{\textbf{RM-Bench}}       \\
\multicolumn{1}{c}{}                                              &                                   &                                          & Chat & Math & Code & Safety & \textit{Avg}  \\ \midrule
\multicolumn{3}{l}{Skywork-RM}                                                                                                                   & 68.7 & 62.0 & 52.8 & 95.9   & 69.9          \\
\quad \textit{w/ CHARM with old Elo}                                                   & 1273                              & 1216                                     & 73.9 & 62.4 & 53.9 & 95.8   & \textbf{71.5} \\
\quad \textit{w/ CHARM with updated Elo}                                                 & 1316                              & 1278                                     & 72.5 & 62.9 & 54.4 & 95.7   & \textbf{71.4} \\ \bottomrule
\end{tabular}

\label{appendix_update_elo}
\end{table*}
\begin{table*}[]
\caption{Comparison of CHARM vs. finetuning on Arena battles. CHARM achieves both improved RM performance and reduced model preference bias, while Arena fine-tuning degrades performance and fails to mitigate bias.}
\centering
\begin{tabular}{@{}lcccccc@{}}
\toprule
\multicolumn{1}{c}{\multirow{2}{*}{\textbf{Models}}} & \multirow{2}{*}{\textbf{\begin{tabular}[c]{@{}c@{}}Mismatch\\ Degree\end{tabular}}} & \multicolumn{5}{c}{\textbf{RM-Bench}}       \\
\multicolumn{1}{c}{}                                 &                                                                                     & Chat & Math & Code & Safety & \textit{Avg}  \\ \midrule
Skywork-RM                                           & 0.639                                                                               & 68.7 & 62.0 & 52.8 & 95.9   & 69.9          \\
\quad \textit{w/ CHARM}                                                & 0.03                                                                                & 73.9 & 62.4 & 53.9 & 95.8   & 71.5 \\
\quad \textit{w/ finetune on} \texttt{arena-55k}                               & 0.459                                                                               & 56.9 & 61.4 & 51.4 & 71.7   & 60.4 \\ \bottomrule
\end{tabular}

\label{appendix_arena_battles}
\end{table*}
\begin{table*}[]
\caption{Multiple models calibration results on RM-Bench.}
\centering
\begin{tabular}{@{}lccccc@{}}
\toprule
\multicolumn{1}{c}{\multirow{2}{*}{\textbf{Method}}} & \multicolumn{5}{c}{\textbf{RM-Bench}}       \\
\multicolumn{1}{c}{}                                             & Chat & Math & Code & Safety & \textit{Avg}  \\ \midrule
Skywork-RM                                                       & 68.7 & 62.0 & 52.8 & 95.9   & 69.9          \\
\quad \textit{w/ CHARM with single model}                                                     & 73.9 & 62.4 & 53.9 & 95.8   & \textbf{71.5} \\
\quad \textit{w/ CHARM with multiple models}                                                   & 73.4 & 63.3 & 54.4 & 95.4   & \textbf{71.6} \\ \bottomrule
\end{tabular}
\label{appendix_multiple}
\end{table*}

Chatbot Arena maintains a dynamic leaderboard where Elo ratings are continuously updated as new models enter and accumulate battle data. A concern is whether \our would require frequent recalibration as these ratings evolve. In this section, we evaluate the robustness of \our with respect to Elo rating updates.

In Equation\ref{eq:elo_definition}, \our relies on Elo differences between model pairs rather than their absolute Elo values. When new models join the Arena, the absolute Elo scores of existing models may shift. However, the relative ranking among existing models tends to remain stable, especially between the strong–weak model pairs that \our targets.

To validate this robustness, we conducted an experiment using Elo scores from two different time periods. One retrieved during our initial experiments and another from the most recent leaderboard. Table~\ref{appendix_update_elo} reports the RM-Bench performance comparison.

Despite substantial Elo shifts (+43 for the reference model and +62 for the over-valued model), the resulting performance difference is negligible. Both sets of Elo scores lead to nearly identical calibration improvements over the baseline. These results demonstrate that \our is robust to updates in the Elo leaderboard.

\subsection{Comparison with Finetuning on Real Arena Battles}

A natural question is whether directly fine-tuning reward models on human
preference data from Chatbot Arena could achieve similar bias mitigation effects as \our. In this section, we investigate this alternative approach and explain why \our is more effective.

Datasets like LMSYS-Chat-1M \citep{zheng2023lmsyschat1m} or arena-human-preference-55k \citep{chiang2024chatbot} contain human preference battles across many models, but they lack sufficient samples involving specific over-valued models we need to correct, especially for smaller or research models (like Gemma-2-9b-it-SimPO we used). Instead \our uses Elo for they aggregate information across many battles and transitively through other models, providing more stable estimates than potentially sparse direct battles.

We conducted an experiment comparing \our against direct fine-tuning on Arena preference data. We finetuned Skywork-RM on arena-human-preference-55k, a dataset of real battles from Chatbot Arena for 1 epoch. Results see Table \ref{appendix_arena_battles}.

From the results we can see finetuning on arena-human-preference-55k leads to a substantial drop in RM-Bench performance and the Mismatch Degree remains high, indicating that general preference data does not address the specific model preference bias.

\subsection{Extend CHARM to Multiple Model Calibration}

In practice, it may be necessary to calibrate more than one over-valued model at the same time. Thus, it is important to see whether \our can generalize to a multi-model setting. In this section, we extend \our to calibrate multiple over-valued models simultaneously.

We select four policy models, \textit{Qwen2.5-72B-Instruct}, \textit{Gemma-2-9b-it-SimPO}, \textit{Llama-3.1-70B-Instruct}, and \textit{GPT-4o-mini-2024-07-18}, and score their responses on Preference20K using Skywork-RM. We jointly optimize offsets for all models so that their calibrated scores match the entire Elo-derived win-rate matrix following steps below:

1. Construct the N × N target win-rate matrix $M_{\text{target}} $ from Elo scores, where

$$P(\textit{i vs. j}) = \displaystyle\frac{1}{1+10^{(Elo_j-Elo_i)/400}}$$

2. Optimize global offsets $\{\Delta_1, \Delta_2, \ldots, \Delta_N\}$ by minimizing

$$\mathbf{MSE}(M_{current},M_{target})$$

3. Construct the calibrated preference dataset using candidate responses from all four models, with calibrated scores $s^\prime_i = s_i+\Delta_i$.

This multi-model extension evaluates whether \our can align an entire group of models with the human preference hierarchy derived from the Elo.

The result in Table~\ref{appendix_multiple} shows multiple model calibration achieves comparable performance to single-model calibration, demonstrating that joint optimization across multiple models maintains effectiveness.

\end{document}